\pdfoutput=1

\documentclass[11pt]{article}
\usepackage[]{acl}

\usepackage{times}
\usepackage{latexsym}

\usepackage[T1]{fontenc}

\usepackage[utf8]{inputenc}

\usepackage{microtype}

\usepackage{soul} %
\usepackage{makecell} %
\usepackage{enumerate}
\usepackage{amsmath}
\usepackage{tikz}
    \usetikzlibrary{decorations.pathreplacing}
    \usetikzlibrary{calc}
    \newcommand{\tikzmark}[1]{\tikz[remember picture,overlay]\node[yshift=2pt](#1){};}
\usepackage[table]{colortbl}
\usepackage{amssymb}
\usepackage{caption}
\usepackage{subcaption}
\usepackage{longtable}
\usepackage{supertabular}
\usepackage{multirow}
\usepackage{pifont}
\newcommand{\cmark}{\ding{51}}%
\newcommand{\xmark}{\ding{55}}%

\usepackage{microtype}
\usepackage{cprotect} %

\usepackage{booktabs}
\usepackage{graphicx}

\title{
Same Author or Just Same Topic? Towards Content-Independent Style Representations
}

\author{Anna Wegmann, Marijn Schraagen and Dong Nguyen  \\
  Department of Information and Computing Sciences \\
  Utrecht University \\
  Utrecht, the Netherlands \\
  \texttt{a.m.wegmann, m.p.schraagen, d.p.nguyen@uu.nl} \\}

\date{}

\begin{document}
\maketitle
\begin{abstract}
Linguistic style is an integral component of language. Recent advances in the development of style representations have increasingly used training objectives from \textit{authorship verification (AV)}: Do two texts have the same author? The assumption underlying the AV training task (same author approximates same writing style) enables self-supervised and, thus, extensive training. However, a good performance on the AV task does not ensure good ``general-purpose'' style representations. For example, as the same author might typically write about certain topics, representations trained on AV might also encode content information instead of style alone. We introduce a variation of the AV training task that controls for content using conversation or domain labels. We evaluate whether known style dimensions are represented and preferred over content information through an original variation to the recently proposed \texttt{STEL} framework. We find that representations trained by controlling for conversation are better than representations trained with domain or no content control at representing style independent from content. 
\end{abstract}

\section{Introduction}

Linguistic style (i.e., how something is said) is an integral part of natural language. 
Style is relevant for natural language understanding and generation \cite{nguyen-etal-2021-learning, ficler-goldberg-2017-controlling} as well as the stylometric analysis of 
texts \cite{el2014authorship, Goswami_Sarkar_Rustagi_2009}. %
Applications include author profiling \cite{10.1145/1871985.1871993} and style preservation in machine translation systems \cite{niu-etal-2017-study, rabinovich-etal-2017-personalized}.

While authors are theoretically able to talk about any topic and (un-)consciously choose to use many styles (e.g., designed to fit an audience  \cite{bell-audience-design}), it is typically assumed that there are combinations of style features that are distinctive for an author (sometimes called an author's {idiolect}).
Based on this assumption, the \textit{authorship verification} task (AV) aims to predict whether two texts have been written by the same author \cite{coulthard2004author, neal_aa-survey, utility_content_AA}. 
Recently, training objectives based on the AV task have been used to train style representations \cite{AV_SimilarityLearning, hay-etal-2020-representation,zhu-jurgens-2021-idiosyncratic}. %
Training objectives on AV are especially promising 
because they do not require any additional labeling when author identifiers are available. Similar to the distributional hypothesis, the assumption underlying the AV training task (same author approximates same writing style) enables extensive self-supervised learning. %

\begin{figure}[t]
    \centering
    \small
   	    \begin{tabular}{p{0cm} p{0.2cm} p{4cm} p{0.2cm}}
            & \color{blue} $A_1$ 
                & \cellcolor{green!25} \textcolor{black}{don’t suggest an open relationship if you’re not ready} \tikzmark{A_end}  \\ 
        \end{tabular}
        
        \vspace{1\baselineskip}
   	    
   	    \begin{tabular}{p{0.2cm} p{2cm} p{0.2cm} p{0.01cm} p{3.5cm} p{0.01cm}}
                \color{blue} $A_2$ \tikzmark{SA_start}
                & \tikzmark{SA_top} \cellcolor{green!25} it’s clear that these are wildly different situations                
                & \tikzmark{SA_end} & \ \ \ \color{blue} $B$  \tikzmark{DA_start} &  \textcolor{gray}{Aren\textquotesingle t open relationships usually just about fixing something in the relationship?}  &  \tikzmark{DA_end} \\ 
            \end{tabular}
            
            \begin{tikzpicture}[overlay, remember picture]
                \draw [decorate,decoration={brace,amplitude=10pt,mirror,raise=4pt}] ($(DA_start.west) - (-0.1,0.3)$) --node[below=14pt]{\color{blue} %
                CC - Same Topic as $A_1$} ($(DA_end.east) - (0.4,1)$);
                \draw [decorate,decoration={brace,amplitude=10pt,mirror,raise=4pt}] ($(SA_start.west) - (-0.1,0.45)$) --node[below=14pt]{\color{blue} Same Author as $A_1$} ($(SA_end.east) - (0.2,1)$);
            \end{tikzpicture}
            
            \vspace{2.5\baselineskip}
    \caption{\textbf{Contrastive Authorship Verification (CAV) Setup and Content Control (CC) Variable.} %
    The CAV task is to match $A_1$ with the utterance $A_2$ that was written by the same author. %
    Contrary to the traditional authorship verification task (AV), this is complemented by a third ``constrastive'' utterance that was written by a different author ($B$). %
    In addition to the CAV variation to AV, we experiment with content control (CC) by selecting $B$ and $A_1$ to have the same approximate content with the help of a topic proxy. As topic proxies we use conversation and domain information. %
    }
    \label{fig:Task}
\end{figure}
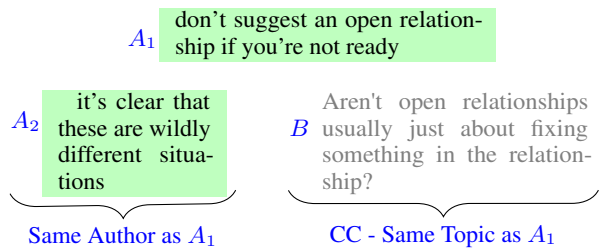

Style and content are often correlated \cite{gero-etal-2019-low,bischoff2020importance}: For example, people might write more formally about their professional career %
but more informally about personal hobbies. %
As a result, style representations  might encode spurious content correlations \cite{poliak-etal-2018-hypothesis}, especially when their AV training objective does {not control for content} \cite{Halvani_AV_bad-topic-control, sundararajan-woodard-2018-represents}. 
Current style representation learning methods either use no or only limited control for content \cite{hay-etal-2020-representation} or use domain labels to approximate topic \cite{AV_SimLearning_Attention}. \newcite{zhu-jurgens-2021-idiosyncratic} work with 24 domain labels (here: product categories) for more than 100k \texttt{Amazon} reviews to improve generalizability. However, using a small set of labels might be too coarse-grained to fully represent and thus control for content. In this paper, we use  ``content'' and ``topic'' to refer to different concepts. We assume same content (fulfilled if two utterances are paraphrases of each other) implies same topic (e.g., two utterances that discuss personal hobbies), while same topic does not necessarily imply same content. %

\textbf{Approach.} 
We introduce two independent variations to the AV task (see Figure \ref{fig:Task}): adding a contrastive sentence (CAV setup) and addressing content correlation with a topic proxy (CC). %
We train several siamese BERT-based neural networks \cite{reimers-gurevych-2019-sentence} to compare style representations learned with the new variations to the AV task. %
We train on utterances from the platform \texttt{Reddit} but %
our approach could be applied to any other conversation dataset as well.
While previous work mostly aimed for learning representations that represent an author's individual style \cite{AV_SimilarityLearning, hay-etal-2020-representation,zhu-jurgens-2021-idiosyncratic}, we aim for general-purpose style representations. As a result, we evaluate the generated representations on (a) whether known style dimensions (e.g. formal vs. informal) are present in the embedding space (Section \ref{sec:eval-STEL}) and preferred over content information (Section \ref{sec:eval-t-a-STEL}) and (b) whether sentences written by the same author are closer to each other even when they have different content (Section \ref{sec:eval-TT}).

\textbf{Contribution.} With this paper, we (a) contribute an extension of the AV task that aims to control for content (CC) with conversation labels, %
(b) introduce a novel variation of the AV setup by adding a contrastive utterance (CAV setup), %
(c) compare style representations trained with different levels of content control (CC) on two task setups (AV and CAV),
(d) introduce a variation of the \texttt{STEL} framework \cite{wegmann-nguyen-2021-capture} to evaluate whether representations prefer content over style information and (e) demonstrate found stylistic features via agglomerative clustering.
We find that representations trained on the conversation topic proxy are better than representations trained with domain or no content control at representing style independent from content. %
Additionally, combining the conversation topic proxy with the CAV setup leads to better results than combining it with the AV setup. We show that our representations are sensitive to stylistic features like punctuation and apostrophe types such as ’ vs. \textquotesingle \  using agglomerative clustering. We hope to further the development of content-controlled style representations. Our code and data are available on GitHub.\footnote{\url{https://github.com/nlpsoc/Style-Embeddings}}

\section{Related Work}

Recently, deep learning approaches have been used in authorship %
verification \cite{shrestha-etal-2017-convolutional, AV_CNN, AV_SimLearning_Attention, AI_SimilarityLearning_Siamese, hay-etal-2020-representation, Hu_AA-Style-Embedding_Triplet, zhu-jurgens-2021-idiosyncratic}.
Training on transformer architectures like BERT has been shown %
to be competitive with other neural as well as non-neural approaches in AV and style representation \cite{zhu-jurgens-2021-idiosyncratic, wegmann-nguyen-2021-capture}. 
AV methods have controlled for content by restricting the feature space to contain ``content-independent'' features like function words or character n-grams \cite{neal_aa-survey, masking-topic_AA, sundararajan-woodard-2018-represents}. 
However, even these features have been shown to not necessarily be content-independent \cite{litvinova_context-stylometric}.

Semantic sentence embeddings are typically trained using supervised or self-supervised learning \cite{reimers-gurevych-2019-sentence}. For supervised learning, models are often trained on manually labelled natural language inference datasets \cite{conneau-etal-2017-supervised}. For self-supervised learning, \textit{contrastive} learning objectives \cite{contrastive-learning} have been increasingly used. Contrastive objectives push semantically distant sentence pairs apart and pull semantically close sentence pairs together. Different strategies for selecting sentence pairs have been used, e.g., same sentences as semantically close %
vs. randomly sampled as semantically distant sentences \cite{giorgi-etal-2021-declutr, gao-etal-2021-simcse}.
\newcite{reimers-gurevych-2019-sentence} also experiment with a \textit{triplet loss},  which pushes an anchor closer to a semantically close sentence and pulls the same anchor apart from a semantically distant sentence. Semantic representations are typically first evaluated on the task that they have been trained on, e.g., binary tasks for binary contrastive objectives and triplet tasks (similar to Figure \ref{fig:Task}) for triplet objectives \cite{reimers-gurevych-2019-sentence}. Semantic representations are often also evaluated on the STS benchmark \cite{cer-etal-2017-semeval} or semantic downstream tasks like semantic search, NLI \cite{bowman-etal-2015-large, williams-etal-2018-broad} or SentEval \cite{conneau-kiela-2018-senteval}.

Typically, objective functions that are known from semantic embedding learning have been used %
\cite{hay-etal-2020-representation, zhu-jurgens-2021-idiosyncratic} with AV training tasks to learn style representations.
\newcite{zhu-jurgens-2021-idiosyncratic} address possible spurious correlations by sampling half of the different and same author utterances from the same and the other half from different domains (e.g., subreddits for \texttt{Reddit}).  %
Style representations are often trained and evaluated on the AV task \cite{AV_SimLearning_Attention, zhu-jurgens-2021-idiosyncratic, bischoff2020importance}.

\section{Style Representation Learning}

We describe the new Contrastive Authorship Verification setup (CAV) and our approach to content control (CC) in Section \ref{sec:task}. Then we describe the generation of training tasks %
(Section \ref{sec:task-generation}) and the hyperparameters for model training (Section \ref{sec:models}).

\subsection{Training Task} \label{sec:task}
The authorship verification (AV) task is the task of predicting whether two texts are written by the same or different authors. In the following, we introduce two independent variations to the AV task: Adding (1) contrastive information with the CAV setup and (2) content control via topic proxies.  

\textbf{CAV setup.} We  
introduce an adaption of the Authorship Verification task --- the Contrastive Authorship Verification setup (CAV, Figure \ref{fig:Task}): Given an anchor utterance $A_1$ and two other utterances $A_2$ and $B$, the task is to identify which of the two sentences were written by the same author as $A_1$. 
Using a contrastive AV setup adds learnable information to the task (namely the contrast between $A_2$ and $B$ w.r.t.~$A_1$) and enables the use of learning objectives that require three input sentences and have been successful in semantic embedding learning \cite{reimers-gurevych-2019-sentence}. We experiment with both CAV and AV setups for style representation learning. In the future, it is also possible to adapt this setup to include several instead of just one contrastive ``negative'' different author utterance (similar to contrastive semantic learning, e.g., in \newcite{gao-etal-2021-simcse}). %
One task with the CAV setup, which consists of three utterances ($A_1$, $A_2$, $B$), can be split up into two AV tasks: ($A_1$, $A_2$) and ($A_1$, $B$). We compare the CAV and AV setups during evaluation (Section \ref{sec:eval}).

\textbf{Content Control (CC).}
Models optimized for AV have been known to make use of semantic information %
\cite{sari-etal-2018-topic, sundararajan-woodard-2018-represents, Stamatatos_av-w-topic} and to perform badly in cross-topic settings \cite{Halvani_AV_bad-topic-control, bischoff2020importance}. 
Recent studies use AV tasks to train style representations and address possible correlations by controlling for domain \cite{zhu-jurgens-2021-idiosyncratic, AV_SimilarityLearning}. %
However, it is unclear to what extent these domain labels are better (or worse) than other ways of controlling for content.
We compare three different levels of content control by approximating content with the help of a topic proxy. We sample the utterance pairs written by different authors ($B$ and $A_1$ for CAV, c.f.~Figure \ref{fig:Task}) (i) from the same \textit{conversation}, (ii) from the same \textit{domain} (e.g., subreddit for \texttt{Reddit} as in \newcite{zhu-jurgens-2021-idiosyncratic}) or (iii) \textit{randomly} (as a baseline, similar to \newcite{hay-etal-2020-representation}). %
Our newly proposed use of the same conversation ``topic proxy'' is inspired by semantic sentence representation learning, where conversations have previously been used as a proxy for semantic information encoded in utterances \cite{yang-etal-2018-learning, liu-etal-2021-dialoguecse}. %
We test to what extent the three different topic proxies are contributing to content-independent style representations during evaluation (Section \ref{sec:eval-t-a-STEL}). %

\begin{table*}[t]
    \small
    \centering
    \begin{tabular}{l l | r r | r | r r | r r | r r}
     \toprule
      &  & \multicolumn{2}{l|}{\ \ \ \ \ \ \ \ \ \ \ \ \textbf{ Setup }} & \multicolumn{1}{l|}{\textbf{Uttterance}} & \multicolumn{2}{l|}{\ \ \ \ \ \textbf{Author}} & \multicolumn{2}{c|}{\textbf{($A_1$, $A_2$)}} & \multicolumn{2}{c}{\textbf{($A_1$, $B$)}} \\
      \textbf{CC level} &  \textbf{Data Split} & \# AV & \# CAV  &   \#     & \# & ma & co & do & co & do \\
    \midrule
     \multirow{3}{*}{\textbf{Conversation}} 
                & train set & $420{,}000$ & $210{,}000$ & $546{,}757$ & $194{,}836$ & $9$ & $0.27$ & $0.56$ & $1.00$ & $1.00$ \\
                & dev set & $90{,}000$ & $45{,}000$ & $116{,}451$ & $41{,}848$ &  $8$ & $0.26$ & $0.55$ & $1.00$ & $1.00$ \\
                & test set & $90{,}000$ & $45{,}000$ & $116{,}621$ & $41{,}902$ &  $8$ & $0.27$ & $0.55$  & $1.00$ & $1.00$ \\
    \midrule
     \multirow{3}{*}{\textbf{Domain}} 
                & train set & $420{,}000$ & $210{,}000$ & $544{,}587$ & $240{,}065$ & $9$ 
                    & \multicolumn{2}{c|}{same pairs} %
                    & $0.01$ & $1.00$ \\
                & dev set & $90{,}000$ & $45{,}000$ & $116{,}490$ & $50{,}939$ & $8$ 
                    & \multicolumn{2}{c|}{as} %
                    & $0.02$ & $1.00$ \\
                & test set & $90{,}000$ & $45{,}000$ & $116{,}586$ & $51{,}182$ & $8$ 
                    & \multicolumn{2}{c|}{conversation}%
                    & $0.02$ & $1.00$ \\
    \midrule
     \multirow{3}{*}{\textbf{No}} 
                & train set & $420{,}000$ & $210{,}000$ & $548{,}082$ & $270{,}079$ & $9$ 
                    & \multicolumn{2}{c|}{same pairs} %
                    &  $0.00$ & $0.01$ \\
                & dev set & $90{,}000$ & $45{,}000$ & $117{,}149$ & $57{,}352$ & $8$ 
                    & \multicolumn{2}{c|}{as}  %
                    & $0.00$ & $0.01$ \\
                & test set & $90{,}000$ & $45{,}000$ & $117{,}434$ & $57{,}726$ & $8$ 
                    & \multicolumn{2}{c|}{conversation} %
                    & $0.00$ & $0.02$ \\
    \bottomrule
    \end{tabular}
    \caption{\textbf{Data Split Statistics.} Per content control (CC) level, 
    we display the number of tasks per setup (\# CAV, \# AV), unique utterances and authors for each split. We also show the maximum number of times an author occurs as $A_1$'s author (ma) and the fraction of same author ($A_1$, $A_2$) and utterance pairs of different authors ($A_1$, $B$) that occur in the same conversation (co) and  domain (do). %
     }
     \label{table:split-statistics}
\end{table*}

\subsection{Task Generation}
\label{sec:task-generation}

We use a 2018 \texttt{Reddit} sample with utterances from 100 active subreddits\footnote{
        \url{https://zissou.infosci.cornell.edu/convokit/datasets/subreddit-corpus/subreddits_small_sample.txt}
    } %
 extracted via \texttt{ConvoKit} \cite{chang-etal-2020-convokit}\footnote{MIT license}.  Per subreddit, we sample 600 conversations with at least 10 posts (which we call utterances). All subreddits are directed at an English audience, which we infer from the subreddit descriptions.
 
\textbf{Generation.} We removed all invalid utterances\cprotect\footnote{Utterance of only spaces, tabs, line breaks or\\of the form: \verb+""+, \verb+" [removed] "+, \verb+"[ removed ]"+,\\\verb+"[removed]"+, \verb+"[ deleted ]"+, \verb+"[deleted]"+, \verb+" [deleted] "+}. Then, we split the set of authors into a non-overlapping 70\% train, 15\% development and 15\% test author split. For each CC level (conversation, domain, no) and each author split, we generated a set of training tasks, i.e., nine sets in total (see Table \ref{table:split-statistics}).

First, we generated the tasks for the train split of the dataset with conversation content control. %
We sampled 210k distinct utterances $A_1$ from the train author split. We use a weighted sampling process to not overrepresent authors that wrote more utterances than others. The maximum time one author wrote $A_1$ is 9 (c.f. ``ma'' in Table \ref{table:split-statistics}). Then, for each utterance $A_1$, we randomly sampled an utterance $B$ that was part of the same conversation as $A_1$ but written by a different author.  
Then, for all 210k ($A_1$, $B$)-pairs, an utterance $A_2$ was sampled randomly from all utterances written by the same author as $A_1$ and for which $A_1 \neq A_2$ holds. We equivalently sampled 45k tasks for the dev and test. 

For the domain and no CC level, we reuse $A_1$ and $A_2$, to keep as many correlating variables constant as possible. Thus, we only resampled 210k utterances $B$ written by a different author from $A_1$ by sampling from the same domain or randomly.

We make sure that each combination of ($A_1$, $A_2$, $B$) occurs only once. Thus there are no repeating CAV tasks.\footnote{Due to the sampling process, there might be same author ($A_1$, $A_2$) pairs that occur twice. However, this remains unlikely due to the high number of authors and utterances. Overall, the share of repeating pairs remains lower than 1\%.} %
However, it is possible that some utterances occur more than once across tasks. In total, we generate 210k train, 45k dev and 45k test tasks for each CC level (see Table \ref{table:split-statistics}), corresponding to a total of 420k, 90k and 90k AV-pairs when splitting the CAV task into (A, SA) and (A, DA) pairs (c.f.~Section \ref{sec:task}).

\begin{table*}[th]
    \small
    \centering
    \begin{tabular}{l r | l l l | l l l} %
     \toprule
         & & \multicolumn{6}{c}{\textbf{\underline{Testing Task}}} \\
         & & \multicolumn{3}{c|}{\textbf{AV}} & \multicolumn{3}{c}{\textbf{CAV}} \\ %
       \multicolumn{2}{c|}{{\textbf{\underline{Training Task}}}} & \multicolumn{1}{c}{\textbf{Conversation}} & \multicolumn{1}{c}{\textbf{Domain}} &  \multicolumn{1}{c|}{\textbf{No}} & \multicolumn{1}{l}{\textbf{Conversation}} &  \multicolumn{1}{c}{\textbf{Domain}} & \multicolumn{1}{l}{\textbf{No}} \\
       \textbf{Setup} & \textbf{CC level} & AUC $\pm\sigma$ & AUC $\pm\sigma$  %
        & AUC $\pm\sigma$ & acc $\pm\sigma$ %
        & acc $\pm\sigma$ & acc $\pm\sigma$ %
       \\
     \midrule
     \multicolumn{2}{c|}{RoBERTa base}
            & $.53$ & $.57$ & $.61$ 
            & $.53$ & $.58$ & $.63$  \\ 
     \midrule   
     \multirow{3}{*}{{\textbf{AV}}}
        & {\textbf{Conversation}}  %
             & $\mathbf{.69}\pm.02$ & ${.70}\pm.02$ & $.71\pm.02$  %
            & $\mathbf{.68}\pm.02$ & ${.69}\pm.02$  & $.70\pm.02$
            \\ 
        & {\textbf{Domain}}  
            & ${.68}\pm.01$ & $\mathbf{.71}\pm.01$ & ${.73}\pm.02$ %
            & ${.67}\pm.01$ & $\mathbf{.70}\pm.01$ & ${.73}\pm.00$  
            \\
        & {\textbf{No}}  
            & $.58\pm.01$ & $.63\pm.02$ & $\mathbf{.79}\pm.00$ %
            & $.59\pm.01$ & $.66\pm.01$ & $\mathbf{.78}\pm.00$ 
            \\
     \midrule
     \multirow{3}{*}{{\textbf{CAV}}}
        & {\textbf{Conversation}} 
            & $\mathbf{.69}\pm.00$ & ${.70}\pm.00$ & ${.71}\pm.00$ %
             & $\mathbf{.68}\pm.00$ & ${.69}\pm.00$ & ${.70}\pm.00$ 
            \\
        & {\textbf{Domain}} %
            & ${.68}\pm.00$ &  ${.70}\pm.00$ & ${.72}\pm.00$  %
            & $\mathbf{.68}\pm.00$ & $\mathbf{.70}\pm.00$ & ${.72}\pm.01$  %
            \\
        & {\textbf{No}} %
            & $.58\pm.00$ & $.63\pm.03$ & $.77\pm.00$  %
            & $.59\pm.00$ & $.65\pm.00$ & ${.77}\pm.00$  
            \\
     \bottomrule
    \end{tabular} %
    \caption{\textbf{Test Results.} Results for 6 different fine-tuned RoBERTa models on the test sets. %
    We display the accuracy of the models for the contrastive authorship verification setup (CAV) and the AUC for the authorship verification task (AV) with different content control approaches (CC). 
    We display the standard deviation ($\sigma$).
    Best performance per column is boldfaced. Models generally outperform others on the CC level they have been trained on. 
    }
    \label{fig:test-task-results}
\end{table*}

\subsection{Training} \label{sec:models}

We use the \texttt{Sentence-Transformers}\footnote{\url{https://sbert.net/}} python library \cite{reimers-gurevych-2019-sentence}\footnote{with Apache License 2.0} to fine-tune several siamese networks %
based on (1) `bert-base-uncased', (2) `bert-base-cased' \cite{devlin-etal-2019-bert} and (3) `roberta-base' \cite{liu2019roberta}. We expect those to perform well based on previous work \cite{zhu-jurgens-2021-idiosyncratic, wegmann-nguyen-2021-capture}. 
We compare using (a) contrastive loss \cite{contrastive-learning} with the AV setup (Section \ref{sec:task}) tasks and (b) triplet loss \cite{reimers-gurevych-2019-sentence} with the CAV setup (Figure \ref{fig:Task}). The binary contrastive loss function uses a pair of sentences as input while the triplet loss expects three input sentences. 
For the loss functions, we experiment with three different values for the margin hyperparameter (i) 0.4, (ii) 0.5, (iii) 0.6.  %
We train with a batch size of 8 over 4 epochs using 10\% of the training data as warm-up steps. We use the Adam optimizer with the default learning rate (0.00002). %
We leave all other parameters as default. We use the BinaryClassificationEvaluator on the AV setup with contrastive loss and the TripletEvaluator on the CAV setup with triplet loss from \texttt{Sentence-Transformers} to select the best model out of the 4 epochs. The BinaryClassificationEvaluator calculates the accuracy of identifying similar and dissimilar sentences, while the TripletEvaluator checks if the distance between A and SA is smaller than the distance between A and DA. We use cosine distance as the distance function.

\section{Evaluation} \label{sec:eval}

We evaluate the learned style representations on the Authorship Verification task (i.e., the training task) in Section \ref{sec:eval-TT}. Then, we evaluate whether models learn to represent known style dimensions via the performance on the \texttt{STEL} framework \cite{wegmann-nguyen-2021-capture} in Section \ref{sec:eval-STEL}. Last, we evaluate representations on their content-independence with an original manipulation of \texttt{STEL} (Section \ref{sec:eval-t-a-STEL}).

\subsection{Authorship Verification} \label{sec:eval-TT}

We display the AV and CAV performance of trained models in Table \ref{fig:test-task-results}. 
On the development sets, RoBERTa models consistently outperformed the cased and uncased BERT models. Also, different margin values only led to small performance differences (Appendix \ref{sec:app_dev-results}). %
Consequently, in Table \ref{fig:test-task-results}, we only display the performance of the six fine-tuned RoBERTa models on the test sets using the three different content controls (CC) and two different task setups (AV 
and CAV setups)
with constant margin values of 0.5.

AV performance is usually calculated with either (i) AUC or (ii) accuracy using a predetermined threshold \cite{zhu-jurgens-2021-idiosyncratic,KestemontEtAl:CLEF-2021}. %
We use cosine similarity to calculate the similarity between sentence representations. Thus, there is no clear constant default threshold to decide between same and different author utterances. A threshold could be fine-tuned on the development set, however for simplicity we use AUC to calculate AV performance instead.%
We use accuracy for the CAV task --- here no threshold is necessary (cosine similarity is calculated between $A_1$, $A_2$ and $A_1$, $B$ and the highest similarity utterance is chosen). 
This makes the performance scores on the test sets less comparable across setups -- however, comparability of the CAV and AV performance scores are limited in any case as the AV vs. CAV setups are fundamentally different. %
Performance scores can be compared across the same column, i.e., within the same AV and CAV setup. 
We aggregate performance with mean and standard deviation for three different random seeds per model parameter combination.\footnote{We used seeds 103-105. 
A total of 5 out of 18 models did not learn. We re-trained those with different seeds.}  %

Overall, the AV \& CAV training task setup (rows in Table \ref{fig:test-task-results}) lead to similar performance on the test sets. As a result, we do not distinguish between them in this section's discussion. Generally, the representations tested on the CC level they were trained on (diagonal) outperform other models that were not trained with the same CC level. For example, representations trained with the conversation CC level, perform better on the test set with the conversation CC than representations trained with the domain or no CC. 

\textbf{Tasks with the conversation label are hardest to solve.} 
For all models, the performance is lowest on the conversation test set and increases on the domain and further on the random test set. This is in line with our assumption that the conversation test set has semantically closer different author utterance ($A_1$, $B$)-pairs that make the AV task harder due to reduced spurious content cues (Section \ref{sec:task}). %

\textbf{Representations trained with the conversation CC might encode less content information.} 
The average performance across the three CC levels is slightly higher for the models trained with domain than conversation CC level and lowest for no CC.  
Across the three test sets with the different CC levels, the standard deviation in performance is biggest for models trained without CC and smallest for models trained with the conversation CC.
Representations trained with domain or no CC might latch on to more semantic features because they are more helpful on the no and domain CC test sets.
Models learned with the conversation CC might in turn learn more content-agnostic representations.
Overall, a representation that performs well on the AV task alone %
might do so by latching on to content (not style) information. As a result, a good AV performance alone might not be indicative of a good representation of style. We further evaluate the quality of style representations and their content-independence in Sections \ref{sec:eval-STEL} and \ref{sec:eval-t-a-STEL}.

\begin{table*}[t]
    \small
    \centering
    \begin{tabular}{p{0.001cm} p{1mm}| p{2.5mm} p{4mm} | l l | l l | l l | l l }
     \toprule
       & &   \multicolumn{2}{c|}{\textbf{all}} & \multicolumn{2}{c|}{\textbf{formal, $\mathbf{n=815}$}} &  \multicolumn{2}{c|}{\textbf{complex, $\mathbf{n=815}$}}  & \multicolumn{2}{c|}{\textbf{nb3r, $\mathbf{n=100}$}} & \multicolumn{2}{c}{\textbf{c'tion, $\mathbf{n=100}$}} \\
      & 
        & \multicolumn{1}{c}{o} & \multicolumn{1}{c|}{o-c} 
        & \ \ \ \ \ \ \ o & \ \ \ \ \ \ o-c 
        & \ \ \ \ \ \ \ o & \ \ \ \ \ \ o-c 
        & \ \ \ \ \ \ \ o & \ \ \ \ \ \ o-c  
        & \ \ \ \ \ \ \ o & \ \ \ \ \ \ o-c \\
      & &  &  & \ acc$\pm\sigma$ & \ acc$\pm\sigma$ & \  acc$\pm\sigma$ & \ acc$\pm\sigma$ & \ acc$\pm\sigma$ & \ acc$\pm\sigma$ & \ acc$\pm\sigma$ & \ acc$\pm\sigma$ \\
     \midrule
        \multicolumn{2}{c|}{{org}}
            & \textbf{.80} & .05 %
            & .83 & .09  %
            & \textbf{.73} & .01  %
            & \textbf{.94} & \textbf{.13}  %
            & \textbf{1.0} & .00  %
            \\        
    \midrule
     \multirow{3}{*}{\textbf{A}}
        & {\textbf{c}}  %
            & .71 & ${.35}$  %
            & ${.83}\pm.02$ & ${.64}\pm.00$ %
            & $.57\pm.02$ & ${.13}\pm.04$ %
            & $.61\pm.02$ & $.04\pm.01$ %
            & ${.91}\pm.10$ & ${.00}\pm.01$ %
            \\         
        & {\textbf{d}} 
            & $.73$ & $.28$ %
            & ${.84}\pm.01$ & ${.56}\pm.04$ %
            & ${.69}\pm.05$ & $.05\pm.02$  %
            & $.61\pm.02$ & $.03\pm.02$  %
            & ${.98}\pm.03$ & $.00\pm.00$  %
            \\  
        & {\textbf{n}} 
            & $.72$ & $.22$ %
            & $\mathbf{.85}\pm.01$ & ${.46}\pm.04$ %
            & $.57\pm.01$ & $.03\pm.01$  %
            & $.62\pm.04$ & $.05\pm.02$  %
            & ${.98}\pm.01$ & $.00\pm.00$  %
            \\  
    \midrule
     \multirow{3}{*}{\parbox{1cm}{\textbf{C}}}
        & {\textbf{c}}
            & $.71$ & $\mathbf{.42}$  %
            & ${.81}\pm.02$ & $\mathbf{.69}\pm.02$ %
            & ${.59}\pm.01$ & $\mathbf{.24}\pm.02$ %
            & ${.65}\pm.09$ & ${.03}\pm.01$ %
            & ${.99}\pm.02$ & $\mathbf{.04}\pm.02$ %
            \\        

        & {\textbf{d}} 
            & $.71$ & $.32$ %
            & ${.82}\pm.01$ & ${.61}\pm.02$ %
            & $.57\pm.01$ & $.12\pm.01$  %
            & $.64\pm.05$ & $.03\pm.01$  %
            & ${.99}\pm.01$ & ${.01}\pm.01$  %
            \\  
        & {\textbf{n}}
            & $.71$ & $.24$ %
            & $\mathbf{.85}\pm.00$ & $.50\pm.02$%
            & ${.56}\pm.01$ & ${.04}\pm.01$  %
            & ${.59}\pm.03$ & ${.06}\pm.01$  %
            & ${.98}\pm.04$ & ${.00}\pm.00$  %
            \\  
     \bottomrule
    \end{tabular}
    \caption{\textbf{\texttt{STEL} and \texttt{STEL}-Or-Content Results.} We display \texttt{STEL} accuracy across 4 style dimensions ($n=$number of instances) for the same RoBERTa models as in Table \ref{fig:test-task-results}: %
    Per task setup (AV - A, CAV - C) and content control level (conversation - c, domain - d, none - n), the performance on the original (o) and the \texttt{STEL}-Or-Content task instances (o-c) are displayed. Per column, the best performance is boldfaced. %
    For the fine-tuned RoBERTa models, performance generally increases on the \texttt{STEL}-Or-Content task compared to the original RoBERTa model (org). 
     }
    \label{table:results-stle-models}
\end{table*}

\begin{figure}[t]
    \centering
    \small
   	    \begin{tabular}{p{45pt} p{60pt} p{1pt} p{60pt}}
              &  \hspace*{20pt} \color{blue} 1 & & \hspace*{20pt} \color{blue}  2   \\
            \color{blue} Anchor (A) 
                & \cellcolor{green!25} r u a fan of them or something? & 
                & \textcolor{black}{Are you one of their fans? \tikzmark{A2_end} } \\ 
            & & \\
                \color{blue} Sentence (S) 
                & \tikzmark{A2_top} \textcolor{gray}{\st{Oh, and also that young physician got an unflattering haircut}}
                & & \cellcolor{green!25} Oh yea and that young dr got a bad haircut  \\ 
            \end{tabular}
            \begin{tikzpicture}[overlay, remember picture]
                \draw [->,very thick, red]($(A2_end.east) - (1.6,0.2)$) -- ($(A2_top.east) - (-2.35,-0.3)$) ;
            \end{tikzpicture}
    \caption{\textbf{\texttt{STEL}-Or-Content Task.} We take the original \texttt{STEL} instances (figure without manipulations) and move A2 to the sentence position with the different style (here: the more formal A2 replaces the more formal S1). These resulting triple tasks %
    can test if a model prefers style over content cues.} 
    \label{tab:t-a-STEL}
\end{figure}
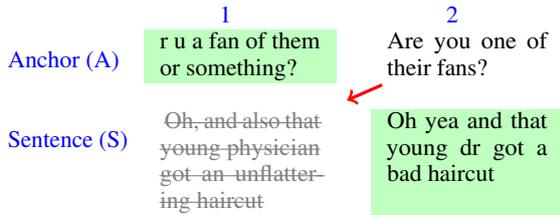

\subsection{STEL Task} \label{sec:eval-STEL}

We calculate the performance of the representations on the \texttt{STEL} framework \cite{wegmann-nguyen-2021-capture}\footnote{\url{https://github.com/nlpsoc/STEL}, with data from \newcite{rao-tetreault-2018-dear, xu-etal-2016-optimizing} and with permission from Yahoo for the ``L6 - Yahoo! Answers Comprehensive Questions and Answers version 1.0 (multi part)'':  \url{https://webscope.sandbox.yahoo.com/catalog.php?datatype=l}. Data and code available with MIT License with exceptions for proprietary Yahoo data.}.
Here, models are evaluated on whether they are able to measure differences in style across four known dimensions of style (formal vs. informal style, complex vs. simple style, contraction usage and number substitution usage). %
Models are tested on 1830 tasks of the same setup: Two ``sentences'' S1 and S2 have to be matched to the style of two given ``anchor'' sentences A1 and A2. The task is binary. Sentences can either be matched without reordering (A1-S1 \& A2-S2) or with reordering (A1-S2 \& A2-S1). For example, consider the sentences in Figure \ref{tab:t-a-STEL} before alterations. The correct solution to the task is to reorder the sentences, i.e., to match A1 with S2 because they both exhibit a more informal style and A2 with S1 because they both exhibit a more formal style. 
The \texttt{STEL} sentence pairs (S1, S2) and (A1, A2) are always paraphrases of each other %
(in contrast to $A_1$ and $B$ for the AV task which are only chosen to be about the same approximate topic, c.f.~\ref{sec:task}). 
The anchor pairs and sentence pairs are randomly matched and are thus otherwise expected to have no connection in content or topic. Representations can thus not make use of learned content features to solve the task. %

We display the \texttt{STEL} results for the RoBERTa models in Table \ref{table:results-stle-models}. 
\textbf{\texttt{STEL} performance is comparable across all fine-tuned models --- for all different CC levels and AV \& CAV setups.} Surprisingly, the overall \texttt{STEL} performance for the fine-tuned models is lower than that of the original RoBERTa base model \cite{liu2019roberta}. Thus, models may have `unlearned' some style information. In the remainder of this subsection, we analyze possible reasons for this \texttt{STEL} performance drop. 

Performance stays approximately the same or improves for the formal/informal and the contraction dimensions, but drops for the complex/simple and the nb3r substitution dimensions. 
Based on manual inspection, we notice nb3r substitution to regularly appear in specific conversations and for specific topics. Future work could investigate whether the use of nb3r substitution is less consistent for one author than other stylistic dimensions. As the nb3r dimension of \texttt{STEL} only consists of 100 instances, future work could increase the number of  instances.  %
Further, we perform an error analysis to investigate the \texttt{STEL} performance drop in the complex/simple dimension. 
We manually look at consistently unlearned (i.e., wrongly predicted by the fine-tuned but correctly predicted by the original RoBERTa model) or learned (i.e., wrongly predicted by the RoBERTa model and correctly predicted by the fine-tuned model) \texttt{STEL} instances (see details in Appendix \ref{sec:app_STEL-error-analysis}). We find several problematic examples where the correct solution to the task is at least ambiguous. We display two such examples in Table \ref{table:err-analysis-unlearned_qualitative}. %
The share of examples with problematic ambiguities is higher for the unlearned (50/55)
than for the newly learned \texttt{STEL} instances (29/41).  Generally, the number of complex/simple \texttt{STEL} instances with ambiguities is surprisingly high for both the learned as well as the unlearned instances, consistent with the lower performance of the models in this category.
Several of the found ambiguities should be relatively easy to correct in the future (e.g., spelling mistakes or punctuation differences).

\subsection{Content-Independence of Style Representations} \label{sec:eval-t-a-STEL}

We tested whether models are able to distinguish between different authors 
(in Section \ref{sec:eval-TT}) and represent styles when the content remains the same
(Section \ref{sec:eval-STEL}). However, we have not tested whether models learn to represent style independent from content. %

Different approaches have been used to test whether style representations encode unwanted content information, including (a) comparing  performance on the AV task across domain \cite{AV_SimilarityLearning, zhu-jurgens-2021-idiosyncratic},
(b) assessing performance on function vs. content words \cite{hay-etal-2020-representation, zhu-jurgens-2021-idiosyncratic} and
(c) predicting domain labels from utterances using their style representations \cite{zhu-jurgens-2021-idiosyncratic}.
However, these evaluation methods have limitations: Domain labels usually come from a small set of coarse-grained labels and function words have been shown to not necessarily be content-independent \cite{litvinova_context-stylometric}. Additionally, next to content, AV might include other spurious features that help increase performance without representing style. %

\begin{table*}[th]
    \centering \small
    \begin{tabular}{c c p{23mm} p{23mm} p{29mm} p{31mm} p{13mm}}
    \toprule
    \textbf{Agg.} & \textbf{GT}  & \textbf{Anchor 1 (A1)} & \textbf{Anchor 2 (A2)} & \textbf{Sentence 1 (S1)}& \textbf{Sentence 2 (S2)} &\textbf{Ambiguity} \\
    \midrule
    \textbf{un} & \cmark 
        &  TDL Group announced in March 2006, in response to a request  [...] %
        & %
            [...] storm names Alberto Helene Beryl Isaac Chris [...]
        & Palestinian voters in the Gaza Strip [...]
            were eligible to participate in the election. 
        &  1. Palestinian voters in the Gaza Strip [...]
            were eligible to participate in the election. 
        & A1/A2 have different content \\
    \midrule
    \textbf{l} 
        & \xmark 
        & [...] %
            51 Phantom [...] %
            received nominations in that same category. 
        & [...] %
            1 phantom [...] %
            received nominations in the same category. 
        & [...] %
            the Port Jackson District Commandant could exchange with all military land with buildings on the harbor. 
        & [...] %
            the Port Jackson District Commandant could communicate with all military installations on the harbour. 
        & A2 spelling mistake, S1 sounds unnatural \\    
    \bottomrule
    \end{tabular}
\caption{\label{table:err-analysis-unlearned_qualitative}
\textbf{\texttt{STEL} Error Analysis.} For the complex/simple \texttt{STEL} dimension, we display examples of ambiguous instances that were learned (l) or unlearned (un) the fine-tuned RoBERTa models. %
A ground truth (GT) of \cmark \ means that S1 matches with A1 and S2 with A2 in style, while \xmark \ means S1 matches with A2 and S2 with A1.
}
\end{table*}

To test if models learn to prefer style over content, we introduce a variation to the \texttt{STEL} framework --- the \textit{\texttt{STEL}-Or-Content} task:
From one original \texttt{STEL} instance (Section \ref{sec:eval-STEL}), %
we take the sentence that has the same style as A2 and replace it with A2. In Figure \ref{tab:t-a-STEL}, this leads to S1 being replaced by A2. The new task is to decide whether A1 matches with the new S1 (originally A2) or with S2. The task is more difficult than the original \texttt{STEL} task as S2 is written in the same style as A1 but has different content and the new S1 is written in a different style but has the same content. The representations will have to decide between giving `style or content' more weight. 
This setup is similar to the CAV task (Figure \ref{fig:Task}). The main differences to the CAV task are (i) that we do not use same author as a proxy for same style but instead use the predefined style dimensions from the \texttt{STEL} framework and (ii) that we control for content with the help of paraphrases (instead of using ony a topic proxy).

We display the \texttt{STEL}-Or-Content results in Table \ref{table:results-stle-models}. The performance for the new task is low ($<0.5$ which corresponds to a random baseline). However, the task is also very difficult as lexical overlap is usually high between the anchor and the false choice (i.e., the sentence that was written in a different style but has the same content). Nevertheless, performance should only be considered in combination with other evaluation approaches (Sections \ref{sec:eval-TT} and \ref{sec:eval-STEL}) as on this task alone models might perform well because they punish same content information. %

\textbf{Models trained on the CAV task with the conversation CC level are the best at representing style independent from content.}
The performance increases from an accuracy of $0.05$ for the original RoBERTa model to up to $0.42\pm.01$ for the representation trained with the CAV task and the conversation CC. %
This `CAV conversation representation' did not just learn to punish same content cues because of its performance on the AV task and the \texttt{STEL} framework: (1) On the AV task, the representation performed comparably on all three test sets. If the model had learned to just punish same content cues, we would expect a clearer difference in performance as confounding same content information should be more prevalent for the random than the conversation test set.
(2) The representation performed comparably to the other representations on the \texttt{STEL} framework, where style information is needed to solve the task but content information cannot be used.

\section{Style Representation Analysis}

We want to further understand 
what the style representations learned to be similar styles. We take the best-performing style representation (RoBERTa trained on the CAV task with the conversation CC and seed 106) and perform agglomerative clustering on a sample of 5.000 CAV tasks of the conversation test set resulting in 14,756 unique utterances. We use 7 clusters based on an analysis of Silhouette scores (Appendix \ref{sec:app_silhouette}). 
Out of all utterance pairs that have the same author, $46.2\%$ appear in the same cluster. %
This is different from random assignments among 7 clusters\footnote{Calculated mean and standard deviation of 100 random assignments of utterances to the 7 clusters of the same size.} which corresponds to $20.1\%\pm .00$. As authors will have a certain variability to their style,
a perfect clustering according to general linguistic style would not assign all same author pairs to the same cluster.

\begin{table}[t]
\renewcommand{\arraystretch}{1.2}
    \small
        \begin{tabular}{c p{14mm} p{47mm}} %
        \toprule
        \textbf{C \#} & \textbf{Consistent} & \textbf{Example}  \\ %
        \midrule
        \textbf{3} %
            & no last punct. & I am living in china, they are experiencing an enormous baby boom 
            \\
        \midrule
        \textbf{4} %
            & 
                punctuation / ca\-sing & huh thats odd i\textquotesingle m in the 97\% percentile on iq tests, the sat, and the  act
            \\
        \midrule
        \textbf{5} %
            & ’ vs \ \textquotesingle 
            & I assume it’s the blind lady? 
            \\
        \midrule
        \textbf{7} %
            & line\-breaks 
            & I admire what you\textquotesingle re doing but [...] %
            \newline  \newline  I know I\textquotesingle m [...]  %
            \\
        \bottomrule
        \end{tabular}
        \caption{\textbf{Clusters for RoBERTa Trained on CAV with Conversation Content Control.} We display one example for 4 out of 7 clusters. 
        We mention noticeable consistencies within the cluster (Consistent). %
        }
        \label{tab:clusters}
\end{table}

In Table \ref{tab:clusters}, we display examples for 4 out of 7 clusters. We manually looked at a few hundred examples per cluster to find consistencies. %
We found clear consistencies within clusters in the punctuation (e.g., 97\% of utterances have no last punctuation mark in Cluster 3 vs. an average of $37\%$ in the other clusters), casing (e.g., $67\%$ of utterances that use \textit{i} instead of \textit{I} appear in Cluster 4), contraction spelling (e.g., 22 out of 27 utterances that use \textit{didnt} instead of \textit{didn't} appear in Cluster 4), the type of apostrophe used (e.g., $90\%$ of utterances use {`} vs {\textquotesingle} \ in Cluster 5 vs. an average of $0\%$ in the other clusters) and line breaks within an utterance (e.g., $72\%$ of utterances in Cluster 7 include line breaks vs. an average of $22\%$ in the other clusters). We mostly found letter-level consistencies --- likely because they are easiest to spot manually. We expect representations to also capture more complex stylometric information because of their performance on the AV and \texttt{STEL} tasks (Section \ref{sec:eval}). Future work could analyze whether and what other stylistic consistencies are represented by the models. %

For comparison we also cluster with the base RoBERTa model (see Appendix \ref{sec:app_cluster}). The only three interesting RoBERTa clusters (i.e., clusters 2,3,4 that contain more than three elements and not as many as $86.7\%$ of all utterances), seem to mostly differ in utterance length (average number of characters are 15 in Cluster 2 vs. in 1278 in Cluster 3) %
and in the presence of hyperlinks ($84\%$ of utterances contain `https://' in Cluster 4 vs. an overall average of $2\%$). %
Average utterance lengths are not as clearly separated by the clusters of the trained style representations. %

\section{Limitations and Future Work}

We propose several directions for future research:

First, conversation labels are already inherently available in conversation corpora like \texttt{Reddit}. However, it remains a difficulty to transfer the conversation CC to other than conversation datasets. 
Moreover, even when using the conversation CC, content information might still be useful for AV: If one person writes ``my husband'' and another writes ``my wife'' within the same conversation, it is highly unlikely that those utterances have been generated by the same person. %
With the recent advances in semantic sentence embeddings, it might be interesting to train style representations on CAV tasks with a new content control level: Two utterances could be labelled as having the same content if their semantic embeddings are close to each other (e.g., when cosine similarity is above a certain threshold).

Second, for the \texttt{STEL}-Or-Content task, the so-called ``triplet problem'' \cite{wegmann-nguyen-2021-capture} remains a potential problem. Consider the example in Figure \ref{tab:t-a-STEL}.
Here, the \texttt{STEL} framework only guarantees that A1 is more informal than A2 and S2 is more informal than S1. Thus, in some cases A2 can be stylistically closer to A1 than S2.  However, we expect this case to be less prevalent: A2 would need to be already pretty close in style to A1, or both S2 and S1 would need to be substantially more informal or formal than A1. %
In the future, removing problematic instances 
could alleviate a possible maximum performance cap.

Third, the representation models may learn to represent individual stylistic variation as we use utterances from the same individual author as positive signals (c.f.~\newcite{zhu-jurgens-2021-idiosyncratic}). However, because the representation models learn with same author pairs that are generated from thousands of authors, it is likely that they also learn consistencies along groups of authors that use similar style features (e.g., demographic groups based on age or education level, or subreddit communities). Future work could explore how different CC levels and training tasks influence the type of styles that are learned.

\section{Conclusion}
Recent advances in the development of style representations have increasingly used training objectives from authorship verification \cite{hay-etal-2020-representation,zhu-jurgens-2021-idiosyncratic}. %
However, representations that perform well on the Authorship Verification (AV) task might do so not because they represent style well but because they latch on to spurious content correlations.
We train different style representations by controlling for content (CC) using conversation or domain membership as a proxy for topic. We also introduce the new Contrastive Authorship Verification setup (CAV) and compare it to the usual AV setup.  %
We propose an original adaptation of the recent \texttt{STEL} framework \cite{wegmann-nguyen-2021-capture} 
to test whether learned representations favor style over content information.
We find that representations that were trained on the CAV setup with conversation CC represent style in a way that is more independent from content than models using other CC levels or the AV setup. %
We demonstrate some of the learned stylistic differences via agglomerative clustering --- e.g., the use of a right single quotation mark vs. an apostrophe in contractions.  
We hope to contribute to increased efforts towards learning general-purpose content-controlled style representations. 

\section*{Ethical Considerations}

We use utterances taken from 100 subcommunities (i.e., subreddits) of the popular online platform \texttt{Reddit} to train style representations with different training tasks and compare their performance. %
With our work, we aim to contribute to the development of general style representations that are disentangled from content. Style representations have the potential to increase classification performance for diverse demographics and social groups \cite{hovy-2015-demographic}.

The user demographics on the selected 100 subreddits are likely skewed towards particular demographics. For example, locally based subreddits (e.g., canada, singapore) might be over-represented. Generally, the average \texttt{Reddit} user is typically more likely to be young and male.\footnote{\small {https://www.journalism.org/2016/02/25/reddit-news-users-more-likely-to-be-male-young-and-digital-in-their-news-preferences/}} 
Thus, our representations might not be representative of (English) language use across different social groups. However, experiments on the set of 100 distinct subreddits should still demonstrate the possibilities of the used approaches and methods.  
We hope the ethical impact of reusing the already published \texttt{Reddit} dataset \cite{Baumgartner_Zannettou_Keegan_Squire_Blackburn_2020, chang-etal-2020-convokit} to be small but acknowledge that reusing it will lead to increased visibility of data that is potentially privacy infringing. 
As we aggregate the styles of thousands of users to calculate style representations, we expect it to not be indicative of individual users. 

We confirm to have read and that we abide by the ACL Code of Ethics.

\section*{Acknowledgements}
We thank the an anonymous ARR reviewers for their helpful comments. This research  was supported by the ``Digital Society - The Informed Citizen'' research programme, which is (partly) financed by the Dutch Research Council (NWO), project 410.19.007. Dong Nguyen was supported by the research programme Veni with project number VI.Veni.192.130, which is (partly) financed by the Dutch Research Council (NWO).

\bibliography{anthology,custom}
\bibliographystyle{acl_natbib}

\clearpage

\appendix

\section{Results on the Development Set}
\label{sec:app_dev-results}

\subsection{Hyperparameter Tuning} \label{sec:hyperparameter}

We evaluated contrastive (on the AV training setup), triple (on the CAV training setup) and online contrastive loss (on the AV training setup) using implementations from \texttt{Sentence-Transformers}. We experiment with the loss hyperparameter ``margin'' with values of 0.4, 0.5, 0.6 for the uncased BERT model \cite{devlin-etal-2019-bert} on the domain training data. Results are displayed in Figure \ref{fig:hyperparameter-tuning}. Contrastive and triplet loss perform better than online contrastive loss. The margin value only has a small influence on the performance scores. Based on these results, we decided to run all further models only with the contrastive and triplet loss functions and a margin value of 0.5.

\begin{table}[ht]
    \small
    \centering
    \begin{tabular}{l | c | l | c | l | c | l } %
     \toprule
        & \multicolumn{2}{c|}{conversation} & \multicolumn{2}{c|}{domain} & \multicolumn{2}{c}{no} \\ %
       & \multicolumn{1}{l|}{CAV} & \multicolumn{1}{c|}{AV} & \multicolumn{1}{l|}{CAV} & \multicolumn{1}{c|}{AV}  & \multicolumn{1}{l|}{CAV} & \multicolumn{1}{c}{AV} \\
       & acc & AUC & acc & AUC & acc & AUC \\
     \midrule   
        {c 0.4}  
            & {0.63} & 0.63 
            & \textbf{0.68} &\textbf{0.68} 
            & \textbf{0.71} & \textbf{0.71} \\
        {c 0.5}  
            & {0.63} & 0.63 
            & \textbf{0.68} &\textbf{0.68} 
            & \textbf{0.71} & \textbf{0.71} \\
        {c 0.6}  
            & {0.62} & 0.63 
            & \textbf{0.68} &\textbf{0.68}
            & \textbf{0.71} & \textbf{0.71} \\ 
        \cmidrule{1-7}
        {t 0.4}  
            & {0.63} & {0.62}
            & {0.68} & {0.67} 
            & 0.70 & 0.70 \\ 
        {t 0.5}  
            & \textbf{0.64} & \textbf{0.64}
            &  \textbf{0.68} &  \textbf{0.68}
            & 0.70 & 0.70 \\
        {t 0.6}  
            & {0.63} & {0.63}
            & {0.67} & {0.67}
            & 0.70 & 0.70 \\ 
        \cmidrule{1-7}
        {c-on 0.4}  
            & {0.58} & 0.58
            & 0.64 & 0.64 
            & 0.67 & 0.67 \\
        {c-on 0.5}  
            & {0.58} & 0.58
            & 0.64 & 0.64
            & 0.67 & 0.67 \\ 
        {c-on 0.6}  
            & {0.58} & 0.58 
            & 0.64 & 0.64 
            & 0.67 & 0.67 \\ 
     \bottomrule
    \end{tabular}
    \caption{\textbf{Hyperparameter-tuning results on the dev AV and CAV datasets with varying content control.} Results for BERT uncased trained on the contrastive authorship verification tasks (CAV). With different loss functions (contrastive - c, triple - t, contrastive online - c-on) and margin values (0.4, 0.5, 0.6). For each dev set (conversation, domain and no content control), we display the accuracy of the models for the CAV task and the AUC for the authorship verification task (AV). For each dev set and CAV/AV setup, the best performance is boldfaced. contrastive and triple loss behave comparable. The margin value only has a small influence.
     }
     \label{fig:hyperparameter-tuning}
\end{table}

\onecolumn
\subsection{Detailed Results on the Development Sets} \label{sec:app_details-dev-results}
We display the performance of further fine-tuned models on the dev sets in Table \ref{fig:task-results}. RoBERTa \cite{liu2019roberta} generally performs better than the uncased and cased BERT model \cite{devlin-etal-2019-bert}. Performance for the triplet and contrastive loss functions are comparable. We only use RoBERTa models in the main paper and both contrastive and triplet loss as a result. 

\begin{table*}[t]
    \small
    \centering
    \subfloat[CAV and AV Performance]{\small
    \begin{tabular}{p{0.01cm} r | p{0.5cm} p{0.5cm} | p{0.5cm}  p{0.5cm} | p{0.5cm}  p{0.5cm}} %
     \toprule
        & & \multicolumn{2}{c|}{conv} & \multicolumn{2}{c|}{sub} & \multicolumn{2}{c}{no} \\ %
       & & \multicolumn{1}{l}{CAV} & \multicolumn{1}{c|}{AV} & \multicolumn{1}{l}{CAV} & \multicolumn{1}{c|}{AV}  & \multicolumn{1}{l}{CAV} & \multicolumn{1}{c}{AV} \\
       & & acc & AUC & acc & AUC & acc & AUC \\
     \midrule
     \multirow{3}{*}{-}
        & bert 
            & $0.52$ & $0.51$  
            & 0.59 & 0.57 
            & 0.64 & 0.61 \\ 
        & BERT 
            & $0.53$ & $0.52$ 
            & 0.59 & 0.57 
            & 0.63 & 0.60 \\
        & RoBERTa 
            & $0.53$ & $0.53$ 
            & $0.58$ & $0.57$ 
            & $0.63$ & $0.61$ \\ 
     \midrule   
     \multirow{7}{*}{\parbox{1cm}{c}}
        & {bert c 0.5} 
            & ${0.65}$ & $0.66$ 
            & $0.66$ & $0.67$ 
            & $0.68$ & $0.68$ \\ 
        & {bert t 0.5} 
            & ${0.65}$ & $0.66$ 
            & $0.66$ & $0.67$ 
            & $0.67$ & $0.68$ \\  
        \cmidrule{2-8}
        & {BERT c 0.5}  
            & ${0.66}$ & ${0.67}$ 
            &  ${0.67}$ &  ${0.68}$ 
            & ${0.69}$ & ${0.70}$ \\ 
        & {BERT t 0.5}  
            & ${0.66}$ & ${0.67}$ 
            & ${0.67}$ & ${0.68}$ 
            & ${0.68}$ & ${0.69}$ \\ 
        \cmidrule{2-8}
        & {RoBERTa c 0.5}  %
            & $\mathbf{0.69}$ & $\mathbf{0.70}$ 
            & $0.70$ & $0.71$ 
            & $0.70$ & $0.72$ \\ 
        & {RoBERTa t 0.5}  
            & ${0.68}$ & ${0.69}$ 
            & ${0.69}$ & ${0.70}$ 
            & ${0.70}$ & ${0.70}$ \\
     \midrule
     \multirow{7}{*}{\parbox{1cm}{s}}
        & {bert c 0.5}  
            & ${0.63}$ & $0.63$ 
            & ${0.68}$ & ${0.68}$  
            & ${0.71}$ & ${0.71}$ \\ 
        & {bert t 0.5}  
            & ${0.64}$ & ${0.64}$
            & ${0.68}$ & ${0.68}$ 
            & $0.70$ & $0.70$ \\
        \cmidrule{2-8} 
        & {BERT t 0.5}  
            & ${0.65}$ & ${0.65}$
            & ${0.68}$ & ${0.68}$ 
            & ${0.71}$ & ${0.71}$ \\ 
        & {BERT c 0.5}  
            & ${0.64}$ & ${0.65}$ 
            & ${0.69}$ & ${0.69}$
            & ${0.71}$ & ${0.72}$ \\ 
        \cmidrule{2-8} 
        & {RoBERTa c 0.5} 
            & ${0.67}$ & ${0.68}$
            & $\mathbf{0.71}$ & $\mathbf{0.72}$
            & ${0.73}$ & ${0.74}$\\
        & {RoBERTa t 0.5} 
            & ${0.68}$ & ${0.68}$
            & ${0.70}$ & ${0.70}$ 
            & ${0.72}$ & ${0.73}$ \\
     \midrule
     \multirow{7}{*}{\parbox{1cm}{r}}
        & {bert c-0.5}  
            & ${0.55}$ & $0.54$ 
            & $0.63$ & $0.62$ 
            & ${0.76}$ & ${0.76}$ \\ 
        & {bert t-0.5}  
            & ${0.55}$ & $0.54$ 
            & $0.62$ & $0.61$ 
            & ${0.74}$ & ${0.75}$ \\ 
        \cmidrule{2-8}
        & {BERT c 0.5}  
            & ${0.57}$ & ${0.56}$ 
            & ${0.64}$ & ${0.63}$ 
            & ${0.76}$ & ${0.77}$  \\
        & {BERT t 0.5}  
            & ${0.58}$ & ${0.56}$ 
            & ${0.64}$ & ${0.62}$ 
            & ${0.75}$ &  {0.75} \\ 
        \cmidrule{2-8} 
        & {RoBERTa c 0.5} 
            & $0.59$ & $0.58$
            & $0.65$ & $0.64$ 
            & $\mathbf{0.77}$ & $\mathbf{0.78}$ \\
        & {RoBERTa t 0.5} 
            & $0.59$ & $0.57$
            & $0.65$ & $0.63$
            & $\mathbf{0.77}$ & $0.77$  \\
     \bottomrule
    \end{tabular}}
     \qquad
    \subfloat[Details on the AV results]{ \small
        \begin{tabular}{c c | c c | c c } %
     \toprule
        \multicolumn{2}{c|}{conv} & \multicolumn{2}{c|}{sub} & \multicolumn{2}{c}{no}  \\ %
      \multicolumn{2}{c|}{AV} & \multicolumn{2}{c|}{AV}  & \multicolumn{2}{c}{AV} \\
      thr & acc & thr & acc & thr & acc \\  %
     \midrule
            $0.82$ & $0.51$ %
            & $0.70$ & $0.55$  %
            & $0.69$ & $0.58$ \\ %
            $0.86$ & $0.51$ %
            & $0.85$ & $0.55$ %
            & $0.85$ & $0.58$ \\ %
            $0.96$ & $0.52$ %
            & $0.97$ & $0.55$ %
            & $0.97$ & $0.58$ \\ %
     \midrule   
            $0.72$ & $0.61$ %
            & $0.73$ & $0.62$ %
            & $0.73$ & $0.63$ \\  %
            $0.27$ & $0.61$  %
            & $0.27$ & $0.62$ %
            & $0.29$ & $0.63$ \\  %
        \cmidrule{1-6} 
            $0.24$ & $0.62$ %
            & $0.28$ & $0.63$  %
            & $0.26$ & $0.64$ \\ %
            $0.72$ & $0.62$ %
            & $0.73$ & $0.63$ %
            & $0.73$ & $0.64$ \\ %
        \cmidrule{1-6}             
            $0.72$ & $\mathbf{0.64}$ %
            & $0.72$ & $0.64$ %
            & $0.73$ & $0.65$ \\ %
            $0.30$ & $0.63$ %
            & $0.31$ & $0.64$ %
            & $0.32$ & $0.64$ \\   
     \midrule
            $0.73$ & $0.59$ %
            & $0.73$ & $0.63$ %
            & $0.73$ & $0.65$ \\ %
            $0.16$ & $0.60$ %
            & $0.19$ & $0.63$ %
            & $0.19$ & $0.64$ \\ %
        \cmidrule{1-6} 
            $0.20$ & $0.61$ %
            & $0.27$ & $0.63$ %
            & $0.23$ & $0.65$ \\ %
            $0.74$ & $0.60$  %
            & $0.74$ & $0.64$  %
            & $0.72$ & $0.66$  \\ %
        \cmidrule{1-6} 
            $0.72$ & $0.63$ %
            & $0.72$ & $\mathbf{0.65}$ %
            & $0.72$ & $0.67$  \\
            $0.22$ & $0.63$  %
            & $0.24$ & $\mathbf{0.65}$  %
            & $0.19$ & $0.66$  \\
     \midrule
            $0.76$ & $0.53$ %
            & $0.77$ & $0.58$ %
            & $0.74$ & $0.69$ \\ %
            $0.14$ & $0.53$  %
            & $0.37$ & $0.57$ %
            & $0.24$ & $0.68$ \\ %
        \cmidrule{1-6} 
            $0.40$ & $0.54$ %
            & $0.35$ & $0.59$ %
            & $0.23$ & $0.69$ \\ %
            $0.74$ & $0.54$  %
            & $0.76$ & $0.59$  %
            & $0.74$ & $0.69$  \\ %
        \cmidrule{1-6} 
            $0.80$ & $0.56$ %
            & $0.77$ & $0.60$ %
            & $0.74$ & $\mathbf{0.71}$  \\
            $0.38$ & $0.55$  %
            & $0.34$ & $0.59$  %
            & $0.19$ & $0.66$  \\
     \bottomrule
    \end{tabular}
    }
    \caption{\textbf{(Dev) Results.} We display the accuracy of the models for the contrastive authorship verification (CAV) setup and the AUC for the authorship verification (AV) setup on each dev set (conversation, domain and no). We show results for 18 fine-tuned models: BERT uncased (bert), RoBERTa and BERT cased  trained with the conversation, domain and no content control. With different loss functions (contrastive - c, triple - t) and margin values (0.4, 0.5, 0.6). For the AV task, we also display the optimal threshold according to AUC (thr) and its matching accuracy. Generally, RoBERTa models perform the best with increasing performance from conversation to domain to random. Accuracies for< CAV are higher than for AV. Models perform the best on the task they have been trained on. Contrastive and Triple loss seem to behave comparable. Best performance per dev set and CAV/AV task is boldfaced. 
     }
     \label{fig:task-results}
\end{table*}

\clearpage

\onecolumn
\section{Details on \texttt{STEL} results}  \label{sec:app_stel-details}
We display the \texttt{STEL} results on further trained models in Table \ref{tab:stel_add}. Interestingly, cased BERT seems to be the better choice for the contraction \texttt{STEL} dimension.

\begin{table*}[t]
    \small
    \centering
    \begin{tabular}{l l| l l | l l | l l | l l | l l }
     \toprule
      train data & model &   \multicolumn{2}{c|}{all} & \multicolumn{2}{c|}{formal} &  \multicolumn{2}{c|}{complex}  & \multicolumn{2}{c|}{nb3r} & \multicolumn{2}{c}{c'tion} \\
      & & \texttt{STEL} & o-c & \texttt{STEL} & o-c & \texttt{STEL} & o-c & \texttt{STEL} & o-c & \texttt{STEL} & o-c \\
     \midrule
     \multirow{1}{*}{-}
        & {BERT uncased (bert)} 
            & 0.75 & 0.03  %
            & 0.76 & 0.05  %
            & 0.70 & 0.00  %
            & \textbf{0.93} & 0.09  %
            & 1.00  & 0.00  %
            \\
        & {BERT cased (BERT)}
            & \textbf{0.78} & 0.05  %
            & 0.80 & 0.10  %
            & \textbf{0.71} & 0.00  %
            & 0.92 & \textbf{0.11} %
            & 1.00  & 0.00   %
            \\        
    \midrule
     \multirow{4}{*}{conv.}
        & {bert c 0.5}
            & 0.68 & {0.21}  %
            & 0.72 & {0.40}  %
            & 0.59 & {0.07}  %
            & 0.73 & {0.06}  %
            & 1.00 & {0.01}  %
            \\   
        & {bert t 0.5}
            & 0.68 & {0.30}  %
            & 0.71 & {0.52}  %
            & 0.61 & {0.15}  %
            & 0.72 & 0.05  %
            & 0.99 & {0.06}  %
            \\   
        \cmidrule{2-12}
        & {BERT c 0.5}
            & 0.73 & 0.32  %
            & 0.83 & 0.62  %
            & 0.60 & \textbf{0.19}  %
            & 0.67 & 0.06  %
            & 1.00 & 0.00  %
            \\     
        & {BERT t 0.5}
            & 0.73 & \textbf{0.37}  %
            & 0.79 & \textbf{0.66}  %
            & 0.63 & 0.15  %
            & 0.74 & 0.05  %
            & 1.00 & \textbf{0.15}  %
            \\     
    \midrule
     \multirow{11}{*}{\parbox{1cm}{domain}}
        & {bert c 0.4}
            & {0.70} & {0.12}  %
            & {0.76} & {0.26}  %
            & {0.61} & {0.01}  %
            & {0.72} & 0.02  %
            & {1.00} & {0.00}  %
            \\      
        & {bert c 0.5}
            & 0.69 & {0.13}  %
            & 0.74 & {0.27}  %
            & 0.59 & {0.01}  %
            & 0.68 & {0.05}  %
            & 1.00 & {0.00}  %
            \\ 
        & {bert c 0.6}
            & {0.70} & {0.13}  %
            & {0.76} & {0.26}  %
            & {0.61} & {0.01}  %
            & {0.72} & 0.04  %
            & {1.00} & {0.00}  %
            \\ 
        & {bert c-on 0.4}
            & 0.65 & {0.02}  %
            & 0.67 & {0.03}  %
            & 0.60 & {0.00}  %
            & 0.69 & 0.02  %
            & 0.84 & {0.00}  %
            \\             
        & {bert c-on 0.5}
            & 0.65 & {0.02}  %
            & 0.67 & {0.03}  %
            & 0.60 & {0.00}  %
            & 0.69 & 0.02  %
            & 0.84 & {0.00}  %
            \\ 
        & {bert c-on 0.6}
            & 0.65 & {0.02}  %
            & 0.67 & {0.03}  %
            & 0.60 & {0.00}  %
            & 0.69 & 0.02  %
            & 0.84 & {0.00}  %
            \\ 
        & {bert t 0.4}
            & {0.71} & {0.15}  %
            & {0.78} & {0.31}  %
            & {0.59} & {0.01}  %
            & {0.78} & 0.05  %
            & 1.00 & {0.00}  %
            \\ 
        & {bert t 0.5}
            & 0.68 & {0.18}  %
            & 0.74 & {0.37}  %
            & 0.58 & {0.03}  %
            & 0.72 & {0.06}  %
            & 1.00 & {0.00}  %
            \\ 
        & {bert t 0.6}
            & 0.69 & {0.22}  %
            & 0.76 & {0.44}  %
            & 0.58 & {0.04}  %
            & 0.69 & 0.06  %
            & 1.00 & {0.00}  %
            \\ 
        \cmidrule{2-12}    
        & {BERT c-0.5}
            & 0.73 & 0.23  %
            & 0.82 & 0.48 %
            & 0.61 & 0.02 %
            & 0.77 & 0.03 %
            & 1.00 & 0.00 %
            \\  
        & {BERT t-0.5}
            & 0.71 & {0.28}  %
            & 0.81 & {0.56}  %
            & 0.57 & {0.06} %
            & 0.80 & 0.04   %
            & 1.00 & 0.00  %
            \\    
    \midrule
     \multirow{4}{*}{random}
        & {bert c 0.5} 
            & 0.69 & 0.09  %
            & 0.77 & 0.20  %
            & 0.58 & 0.01  %
            & 0.68 & 0.02  %
            & 0.98 & 0.00  %
            \\
        & {bert t 0.5} 
            & 0.70 & 0.13  %
            & 0.75 & 0.26  %
            & 0.61 & 0.03  %
            & 0.79 & 0.06  %
            & 1.00 & 0.00  %
            \\
        \cmidrule{2-12}    
        & {BERT c-0.5} 
            & 0.72 & 0.21  %
            & \textbf{0.84} & 0.44 %
            & 0.55 & 0.02 %
            & 0.75 & 0.07 %
            & 1.00 & 0.01 %
            \\  
        & {BERT t-0.5}
            & 0.73 & {0.23}  %
            & \textbf{0.84} & {0.48}  %
            & 0.59 & {0.03} %
            & 0.68 & 0.05   %
            & 1.00 & 0.00  %
            \\    
     \bottomrule
    \end{tabular}
    \caption{\textbf{Results on \texttt{STEL} and \texttt{STEL}-Or-Content.} We display \texttt{STEL} accuracy for different language models and methods. %
    The performance on the set of \texttt{STEL}  and \texttt{STEL}-Or-Content (o-c) task instances is displayed. The best performance is boldfaced. Performance for the trained models goes down for the original \texttt{STEL} framework in the complex/simple and nb3r substitution dimension. Performance generally increases for the \texttt{STEL}-Or-Content task.
     }
     \label{tab:stel_add}
\end{table*}

\twocolumn
\subsection{Error Analysis RoBERTa \texttt{STEL} results} \label{sec:app_STEL-error-analysis}

In Table \ref{fig:error-analysis_quantitative}, we display the number of learned and unlearned \texttt{STEL} instances across different aggregates for the RoBERTa models. We combine all such unique \texttt{STEL} instances across the aggregates and annotate if they contain ambiguities. In Table~\ref{tab:error-analysis_categories}, we display the results. Overall, the learned \texttt{STEL} instances contain fewer ambiguities. However, they still show considerable amounts of ambiguities.

\begin{table}[t]
    \small
    \centering
    \begin{tabular}{l l | r r | r r }
     \toprule
      \multicolumn{2}{c|}{aggregate} & \multicolumn{2}{c|}{ unlearned} & \multicolumn{2}{c}{learned} \\
      & & f/i & c/s %
      & f/i & c/s \\ %
    \midrule
        \multirow{3}{*}{CC} 
            & conversation 
                & 21 & 34 %
                & 62 & 22 \\ %
            & domain
                & 13 & 34 %
                & 62 & 24 \\ %
            & no 
                & 21 & 44 %
                & 67 & 24 \\ %
    \midrule
        \multirow{2}{*}{setup} 
            & AV 
                & 8 & 9 %
                & 61 & 11 \\ %
            & CAV 
                & 6 & 14 %
                & 55 & 14 \\ %
    \midrule
        \multirow{1}{*}{-} 
            & all 
                &  1 & 4 %
                & 48 & 8 \\ %
     \bottomrule
    \end{tabular}
    \caption{\textbf{Error Analysis \texttt{STEL} Results.}  For the formal/informal (f/i) and complex/simple (c/s) \texttt{STEL} dimension, we display the number of instances that were unlearned and learned by all RoBERTa models in an aggregate. We use three different aggregates: (i) all models trained with a given CC level, (ii) all models trained with a certain task setup and (iii) all models. 
     }
     \label{fig:error-analysis_quantitative}
\end{table}

\begin{table}[t]
    \small
    \centering
    \begin{tabular}{l | r | r }
     \toprule
       & {unlearned} & {learned} \\
    \midrule
        no ambiguity 
            & $\frac{5}{55}\approx9\%$  %
            & $\frac{12}{41}\approx29\% $ \\  %
        \midrule
        typo simple  
            & $\frac{21}{55}\approx38\%$   %
            & $\frac{13}{41}\approx32\%$ \\  %
        typo complex  
            & $\frac{11}{55}\approx20\%$  %
            & $\frac{6}{41}\approx15\%$ \\  %
        error grammar simple  
            & $\frac{15}{55}\approx27\%$ %
            & $\frac{9}{41}\approx22\%$ \\  %
        error grammar complex  
            & $\frac{5}{55}\approx9\%$ 
            & $\frac{3}{41}\approx7\%$ \\
        \midrule
        changed content  
            & $\frac{5}{55}\approx9\%$ 
            & $\frac{3}{41}\approx7\%$ \\
        \midrule
        word as/more complex  %
            & $\frac{16}{55}\approx29\%$  
            & $\frac{11}{41}\approx27\%$ \\  %
        naturalness 
             & $\frac{7}{55}\approx13\%$
            &  $\frac{3}{41}\approx7\%$ \\
     \bottomrule
    \end{tabular}
    \caption{\textbf{Categories Error Analysis \texttt{STEL} Results.} For the six fine-tuned RoBERTa models, we manually looked at the common learned as well as the unlearned simple/complex examples. We put the examples in the displayed ambiguity classes. 
     }
     \label{tab:error-analysis_categories}
\end{table}

\section{Details on cluster parameters} \label{sec:app_silhouette}

We use agglomerative clustering for the RoBERTa model trained on the CAV setup with a margin of 0.5 and conversations as CC with seed 106 (R CAV CONV 106).  %
We experiment with different numbers of clusters and display the results in Table~\ref{tab:cluster_hyperparameter-tuning}. The highest Silhouette scores are reached for cluster sizes of 5, 6, 7. We select a cluster size of 7 for evaluation. %

\begin{table}[t]
    \small
    \centering
    \begin{tabular}{l | r | r } 
     \toprule
        & n & avg. silhouette \\
     \midrule
        \multirow{30}{*}{} 
         & {2}  & 0.23 \\
         & 3 & 0.21 \\
         & 4 & 0.23 \\
         & \textbf{5} & \textbf{0.27} \\
         & \textbf{6} & \textbf{0.27} \\ 
         & 7 & 0.26 \\
         & 8 & 0.23 \\
         & 9 & 0.19 \\
         & 10 & 0.20 \\
         & 11 & 0.19 \\
         & 12 & 0.18 \\
         & 13 & 0.19 \\
         & 14 & 0.17 \\
         & 15 & 0.16 \\
         & 16 & 0.16 \\
         & 17 & 0.16 \\
         & 18 & 0.17 \\
         & 19 & 0.17 \\
         & 20 & 0.17 \\ %
         & 21 & 0.16 \\
         & 22 & 0.16 \\
         & 23 & 0.15 \\
         & 24 & 0.15 \\
         & {25} & 0.15 \\ %
         & 26 & 0.15 \\
         & 30 & 0.15 \\
         & 40 & 0.15 \\
         & 50 & 0.15 \\
         & 100 & 0.13 \\
         & 150 & 0.13 \\
         & 200 & 0.12 \\
     \bottomrule
    \end{tabular}
    \caption{\textbf{Silhouette values.} We experiment with different numbers of clusters for one fine-tuned RoBERTa model (R CAV CONV 106). It was on the CAV task with conversation CC. The highest Silhouette score is reached for cluster sizes of 5--7.
     }
     \label{tab:cluster_hyperparameter-tuning}
\end{table}

\section{Details on the cluster analysis} \label{sec:app_cluster}

We give more examples of the seven clusters in Table \ref{table:app_cl-examples}. Refer to our Github repository for the complete clustering. We did not find obvious consistencies for clusters 1, 2 and 6. That does, however, not mean that more nuanced stylistic consistencies are not present. We recommend using a higher number of clusters, possibly different clustering algorithms and testing out statistics for known style features to pinpoint more consistencies.

Out of all utterance pairs that have the same author, $46.2\%$ appear in the same cluster for the style embedding model. %
This is different from a random distribution among 7 clusters\footnote{Calculated mean and standard deviation of 100 random assignments of utterances to the 7 clusters, with the same number of elements in each cluster.} which corresponds to $20.1\%\pm.00$. As authors will have a certain variability to their style as well (e.g., \newcite{zhu-jurgens-2021-idiosyncratic}), a perfect clustering according to writing style would not assign all same author pairs to the same cluster.
For the RoBERTa base model the fraction of same author pairs in the same cluster is closer to the random distribution ($75.4\%$ vs. $76.1\%$ %
for the random distribution\footnote{The share is high for RoBERTa base because the first cluster already contains  $86.7\%$ of all utterances.}). 
The fraction of utterance pairs that appear in the same domain are close to the random distribution for both the style embedding model ($23.6\%$ vs. $20.1\%$) and the RoBERTa base model ($77.6\%$ vs. $76.0\%$). The percentage for the RoBERTa base models is a lot higher as the first cluster contains almost $90\%$ of all utterances. Random assignment of utterances across the 7 clusters, that keeps the clustering size would already lead to $76.0\%$ same author pairs appearing in the same cluster (almost all of them in the first).
Results are similar for utterance pairs that appear in the same conversation. 

\begin{table*}[th!]
    \scriptsize
        \begin{tabular}{p{.1mm} c | p{13mm} | p{38mm} p{38mm} p{39mm}}
        \toprule
        \textbf{C} & \textbf{\#}  & \textbf{Consistency} & \textbf{Example 1} & \textbf{Example 2} & \textbf{Example 3} \\
        \midrule
        \textbf{1} & 4065 
            & %
                citing previous comments, standard punctuation, URLs %
            & Yes. Proportionally, this kid's feet are absolutely enormous.
            & > Please delete your account. \newline \newline \newline Says the no life who always shits on anything Kanye or anti-Drake
            I can promise you that capitalism is very much alive in Norway. 
            & [This should help.](YOUTUBE-LINK) \\ %
        \midrule 
        \textbf{2} & 4016 & short sentences?
            & Nice catch! Well done. cookies are in the back of this Grammar party. You can have two. & You can mute them we've been told!
            & Came here to post this only to find it's already the top voted comment. This is a good sub. \\
        \midrule
        \textbf{3} & 2165 
            & no last punctuation mark & I am living in china, they are experiencing an enormous baby boom 
            & Seems like sarcasm. But could also be Poe
            & [...] %
            The earth probably has two or more degrees of symmetry, but less than infinite (like a sphere), but I\textquotesingle m honestly not too concerned about the minutiae of it
            \\
        \midrule
        \textbf{4} & 1794 
            & 
                punctuation / ca\-sing & huh thats odd i\textquotesingle m in the 97\% percentile on iq tests, the sat, and the  act
            & Its not a problem if you a got a full game. Whats the problem if a game didnt get expansions?
            & Fair point, I didnt know that. Just at glance I kind of went \textquotesingle woah that doesnt seem right\textquotesingle
            \\
        \midrule
        \textbf{5} & 1555 & ’ instead of \ \textquotesingle \  apostrophe 
            & I assume it’s the blind lady? 
            & Oh I wasn’t really dismissing them. I’m saying Ford will try their own thing compared to Fiat
            & It’s 4am in Brussels and I am still hyped \\
        \midrule
        \textbf{6} & 781 &  similar to 1?
            & Well, as your neighbors, I'd say Fuck you..  But we're not like that, see?  We want to be part of the alliance, not part of the 'fuck you, we cant be competitive with jobs or innovate any more, so we're going to run massive tariffs against all our friendly nations 
            & Hah, thus the one calf larger than the other issue.  I have it too ;)
            & [So you are saying that current encryption falls apart as long as the quantum computer is large enough](URL). %
            (for reference, the current highest qubit is 50)'
            \\
        \midrule
        \textbf{7} & 380 & line\-breaks 
            & I admire what you\textquotesingle re doing but [...] %
            \newline  \newline  I know I\textquotesingle m in the minority. [...] %
            & 75\% of the problems I run into are solved by [...] %
            \newline  \newline I work in live streaming. 
            & All the suggestions others have given are excellent. RS7 makes the most sense to me.  \newline  \newline But [...] %
            \newline  \newline Meanwhile, [...] %
            \\
        \bottomrule
        \end{tabular}
        \caption{\label{table:app_cl-examples}
        \textbf{Clustering - fined-tuned RoBERTa model.} We display examples for each cluster of the 7 clusters that resulted from the agglomerative clustering of 14,756 randomly sampled texts with the RoBERTa model fine-tuned on the CAV setup with the conversation CC. We mention noticeable consistencies (Consistency) within the cluster and give three examples each. Consistencies that are not as clear are marked with a `?'.}

\end{table*}

\begin{table*}[t]
\renewcommand{\arraystretch}{1.2}
    \scriptsize
        \begin{tabular}{p{.1mm} c | p{13mm} | p{33mm} p{35mm} p{55mm}}
        \toprule
        \textbf{C} & \textbf{\#}  & \textbf{Consistency} & \textbf{Example 1} & \textbf{Example 2} & \textbf{Example 3} \\
        \midrule
        \textbf{1} & 12798 
            & wide variety 
            & Just googled it, looks like a great device for the price! If I weren't so impatient I would have bought this online. Great battery life!
            & This is exactly why i believe iphone 5 body was perfect example of good balance with design(timeless) and utility
            & [...] \newline 
            The earth probably has two or more degrees of symmetry, but less than infinite (like a sphere), but I\textquotesingle m honestly not too concerned about the minutiae of it\\
        \midrule
        \textbf{2} & 1110 
            & short utterances 
            & here we go!! %
            & And her good posture. %
            & Not in California. \\ %
        \midrule
        \textbf{3} & 310 
            & long utterances 
            & I’ve never had the pleasure of seeing Neil live but I got on a big kick a few years ago after buying one of his live albums (can’t remember which one) where I listened to all his live albums and then wanted to see as many of his live performance I could find on YouTube. [...] %
            & \&gt; but the movie has the superior ending I think. \newline \newline 
            [...] \newline \newline
            [...]

            & So .... heavily influenced by the social economics ... but still voluntary, got it. [...] %
            \newline \newline 
            Then how about this. [...] %
            \newline 
            Everyone still keeps their child that way, you even promote child birth. No sterilization, no stigmatization of poor people, no poor people stuck with child with heavy needs requiring care that they can’t pay for. 
            \\
        \midrule
        \textbf{4} & 232 & URLs %
            & \url{https://youtu.be/GmULc5VANsw}
            & [This](\url{https://np.reddit.com/r/MakeupAddiction/comments/25hkqi/how_to_tell_if_your_foundationprimer_is_silicone/}) might help!%
            & I thought there was 51 stars because of Puerto Rico \newline \newline \url{https://en.m.wikipedia.org/wiki/51st_state} \\
        \bottomrule
        \end{tabular}
        \caption{\label{table:quad-examples}
        \textbf{Clusters for RoBERTa base.} We display examples for 4 out of 7 clusters as a result of the agglomerative clustering of 14756 randomly sampled texts from the conversation test set. We mention noticeable consistencies (Consistency) within the cluster and give three examples each.}%
\end{table*}

\clearpage

\section{Computing Infrastructure}
The training of 23 RoBERTa \cite{liu2019roberta}, 13 uncased BERT and 6 cased BERT models \cite{devlin-etal-2019-bert} took about 846 GPU hours  with one RTX6000 card with 24 GB RAM on a Linux computing cluster. Further analysis and clustering of two RoBERTa models took about 24 GPU hours. We used a machine with 32 GB RAM and 8 intel i7 CPUs using Ubuntu 20.04 LTS without GPU access to generate the training data.

We used \texttt{Sentence-Transformers} 2.1.0 \cite{reimers-gurevych-2019-sentence} and \texttt{numpy} 1.18.5 \cite{harris2020array}, \texttt{scipy} 1.5.2 \cite{2020SciPy-NMeth} and \texttt{scikit-learn} 0.24.2 \cite{scikit-learn}.

We use previous work, including code and data, consistent with their specified or implied intended use \cite{reimers-gurevych-2019-sentence, chang-etal-2020-convokit, wegmann-nguyen-2021-capture}. The \texttt{ConvoKit} open-source Python framework invites NLP researchers and `anyone with questions about conversations' to use it \cite{chang-etal-2020-convokit}. The \texttt{SentenceTransformers} Python framework can be used to compute sentence / text embeddings.\footnote{\url{https://sbert.net/}} We comply with asking permission for part of the dataset for \texttt{STEL} and citing the specified works  \cite{wegmann-nguyen-2021-capture}. \newcite{wegmann-nguyen-2021-capture} state the intended use of developing improved style(-sensitive) measures. 

\section{Intended Use} We hope our work will inform further research into style and its representations. We invite researchers to reuse any of our provided results, code and data for this purpose.

\end{document}